\documentclass[10pt,twocolumn,letterpaper]{article}

\usepackage{cvpr}
\usepackage{times}
\usepackage{epsfig}
\usepackage{graphicx}
\usepackage{amsmath}
\usepackage{amssymb}
\usepackage{enumitem}
\usepackage{colortbl}
\usepackage{arydshln}
\usepackage{multirow}
\usepackage{multicol}
\usepackage[toc,page]{appendix}
\usepackage{cite}

\usepackage[pagebackref=true,breaklinks=true,letterpaper=true,colorlinks,bookmarks=false]{hyperref}

\cvprfinalcopy 

\def\cvprPaperID{1936} 

\ifcvprfinal\fi
\begin{document}

\title{Deep Stacked Hierarchical Multi-patch Network for Image Deblurring}

\author{Hongguang Zhang\textsuperscript{1,2,4},\quad Yuchao Dai\textsuperscript{3},\quad Hongdong Li\textsuperscript{1,4}, \quad Piotr Koniusz\textsuperscript{2,1}\\
$^1$Australian National University, \quad$^2$Data61/CSIRO  \\
$^3$Northwestern Polytechnical University, \quad $^4$ Australian Centre for Robotic Vision\\
firstname.lastname@\{anu.edu.au\textsuperscript{1}, data61.csiro.au\textsuperscript{2}\}, daiyuchao@nwpu.edu.cn\textsuperscript{3}
}
\maketitle
\thispagestyle{empty}

\begin{abstract}
Despite deep end-to-end learning  methods have shown their superiority in removing non-uniform motion blur, there still exist major challenges with the current multi-scale and scale-recurrent models: 1) Deconvolution/upsampling operations in the coarse-to-fine scheme result in expensive runtime; 2) Simply increasing the model depth with finer-scale levels cannot improve the quality of deblurring. To tackle the above problems, we present a deep {hierarchical multi-patch network} inspired by Spatial Pyramid Matching to deal with blurry images via a fine-to-coarse hierarchical representation. To deal with the performance saturation w.r.t. depth, we propose a stacked version of our multi-patch model. Our proposed basic multi-patch model achieves the state-of-the-art performance on the GoPro dataset while enjoying a 40$\times$ faster runtime compared to current multi-scale methods. With 30ms to process an image at 1280$\times$720 resolution, it is the first real-time deep motion deblurring model for 720p images at 30fps. For stacked networks, significant improvements (over 1.2dB) are achieved on the GoPro dataset by increasing the network depth. Moreover, by varying the depth of the stacked model, one can adapt the performance and runtime of the same network for different application scenarios. 
\end{abstract}
\vspace{-0.3cm}
\section{Introduction}
The goal of non-uniform blind image deblurring is to remove the undesired  blur caused by the camera motion and the scene dynamics  \cite{nah2017deep,tao2018scale,pan2017simultaneous}. 
Prior to the success of deep learning, conventional  deblurring methods used to employ a variety of constraints or regularizations to approximate the motion blur filters, involving an expensive non-convex nonlinear optimization. Moreover, the commonly used assumption of spatially-uniform blur kernel is overly restrictive, resulting in a poor deblurring of complex blur patterns.

\begin{figure}[t]
\vspace{-0.5cm}
	\centering
	\includegraphics[height=4.5cm]{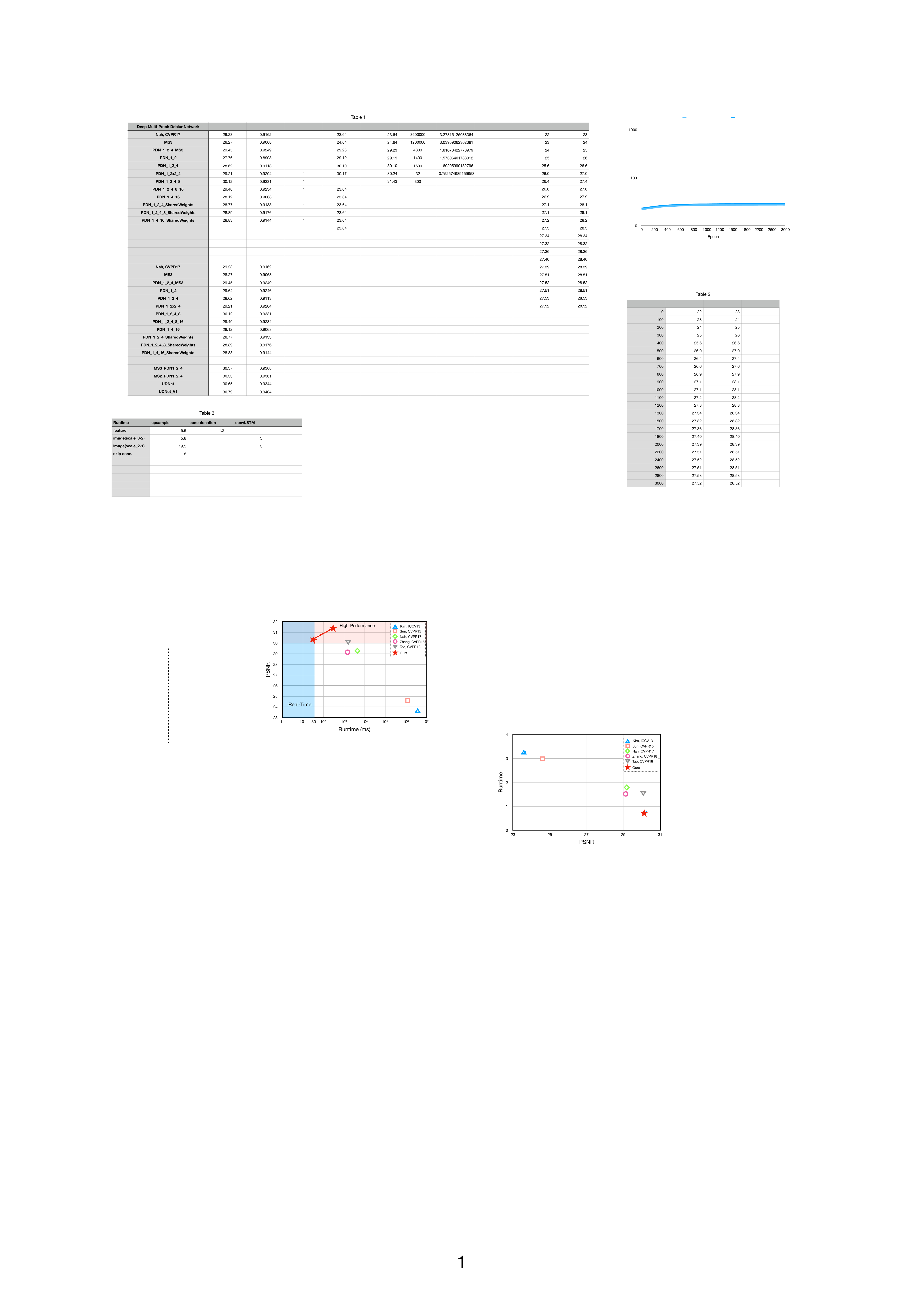}
	\caption{\small The PSNR vs. runtime of state-of-the-art deep learning motion deblurring methods and our method on the GoPro dataset \cite{nah2017deep}. The blue region indicates real-time inference, while the red region represents high performance motion deblurring (over 30 dB). Clearly, our method achieves the best performance at 30 fps for $1280\!\times\!720$ images, which is 40$\times$ faster than the very recent method \cite{tao2018scale}. A stacked version of our model further improves the performance at a cost of somewhat increased runtime.}
	\label{fig:PRPlot}
    \vspace{-0.3cm}
\end{figure}

\begin{figure*}[htp]
\vspace{-0.8cm}
	\centering
	\includegraphics[width=\linewidth]{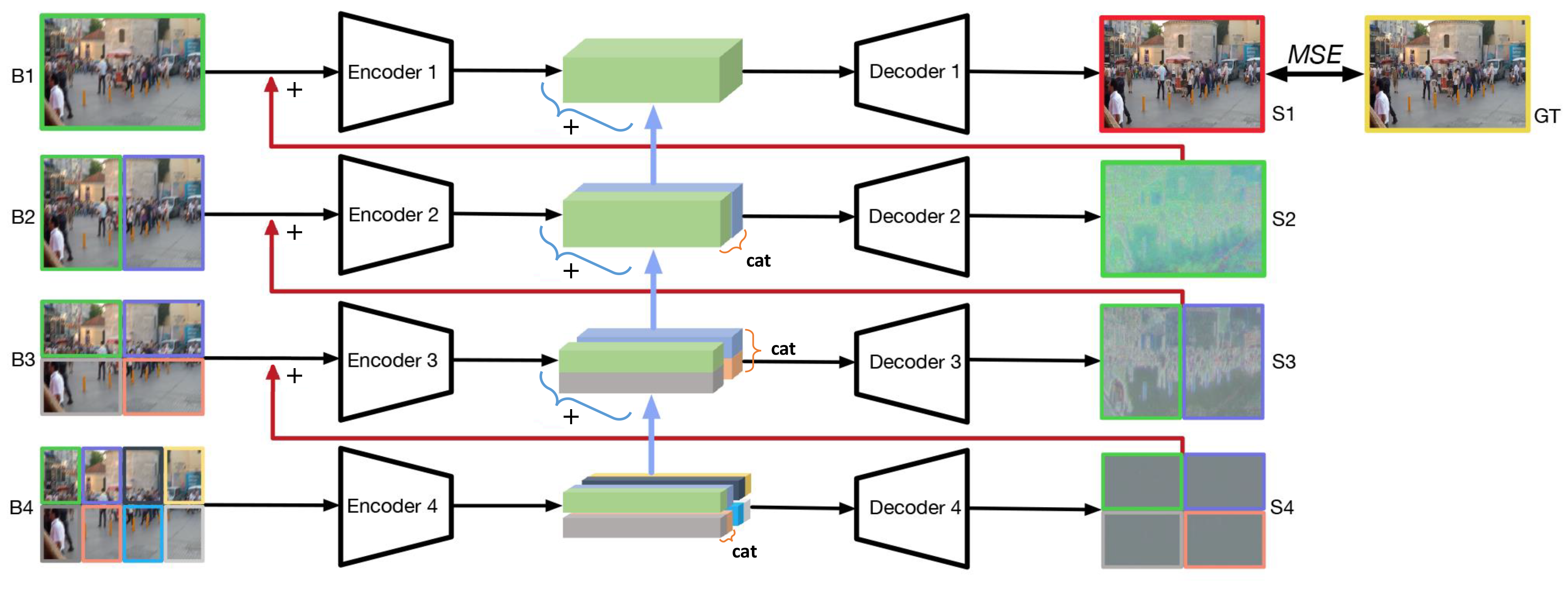}
	\caption{\small Our proposed Deep Multi-Patch Hierarchical Network (DMPHN). As the patches do not overlap with each other,  they may cause boundary artifacts which are removed by the consecutive upper levels of our model. Symbol $+$ is a summation akin to residual networks.}
	\label{fig:DMPHN}
    \vspace{-0.3cm}
\end{figure*}

Deblurring methods based on Deep Convolutional Neural Networks (CNNs) \cite{krizhevsky2012imagenet,simonyan2014very} learn the regression between a blurry input image and the corresponding sharp image in an end-to-end manner \cite{nah2017deep,tao2018scale}.
To exploit the deblurring cues at different processing levels, the ``coarse-to-fine'' scheme has been extended to deep CNN scenarios by  a multi-scale network architecture \cite{nah2017deep} and a scale-recurrent architecture \cite{tao2018scale}. Under the ``coarse-to-fine'' scheme, a sharp image is gradually restored at different resolutions in a pyramid.
Nah \etal \cite{nah2017deep} demonstrated the ability of CNN models to remove motion blur from multi-scale blurry images, where a multi-scale loss function is devised to mimic conventional coarse-to-fine approaches. 
Following a similar pipeline, Tao \etal \cite{tao2018scale} share network weights across scales to improve training and model stability, thus achieving highly effective deblurring compared with \cite{nah2017deep}. 
However, there still exist major challenges in deep deblurring:
\vspace{-0.2cm}
\begin{itemize}
    \item Under the coarse-to-fine scheme, most networks use a large number of training parameters due to large filter sizes. Thus, the multi-scale and scale-recurrent methods result in an expensive runtime (see  Fig.~\ref{fig:PRPlot}) and struggle to improve deblurring quality.
    \vspace{-2mm}
    \item Increasing the network depth for very low-resolution input in multi-scale approaches  does not seem to improve the deblurring performance \cite{nah2017deep}. 
\end{itemize}

In this paper, we address the above challenges with the multi-scale and scale-recurrent architectures. We investigate a new scheme which exploits the deblurring cues at different scales via a \emph{hierarchical multi-patch} model.  Specifically, we propose a simple yet effective multi-level CNN model called Deep Multi-Patch Hierarchical Network (DMPHN) which uses multi-patch hierarchy as input. In this way, the residual cues from deblurring local regions are passed via residual-like links to the next level of network dealing with coarser regions.  Feature aggregation over multiple patches has been  used in image classification \cite{lazebnik2006beyond,he2014spatial,lu2015deep,koniusz2018deeper}. For example, \cite{lazebnik2006beyond} proposes Spatial Pyramid Matching (SPM) which divides images into coarse-to-fine grids in which histograms of features are computed. In \cite{koniusz2018deeper}, a second-order fine-grained image classification model uses overlapping patches for aggregation. Sun \etal \cite{sun2015learning} learned a patch-wise motion blur kernel through a CNN for restoration via an expensive energy optimization.

\iftrue
\begin{figure}[b]
\vspace{-0.3cm}
	\includegraphics[width=1.0\linewidth]{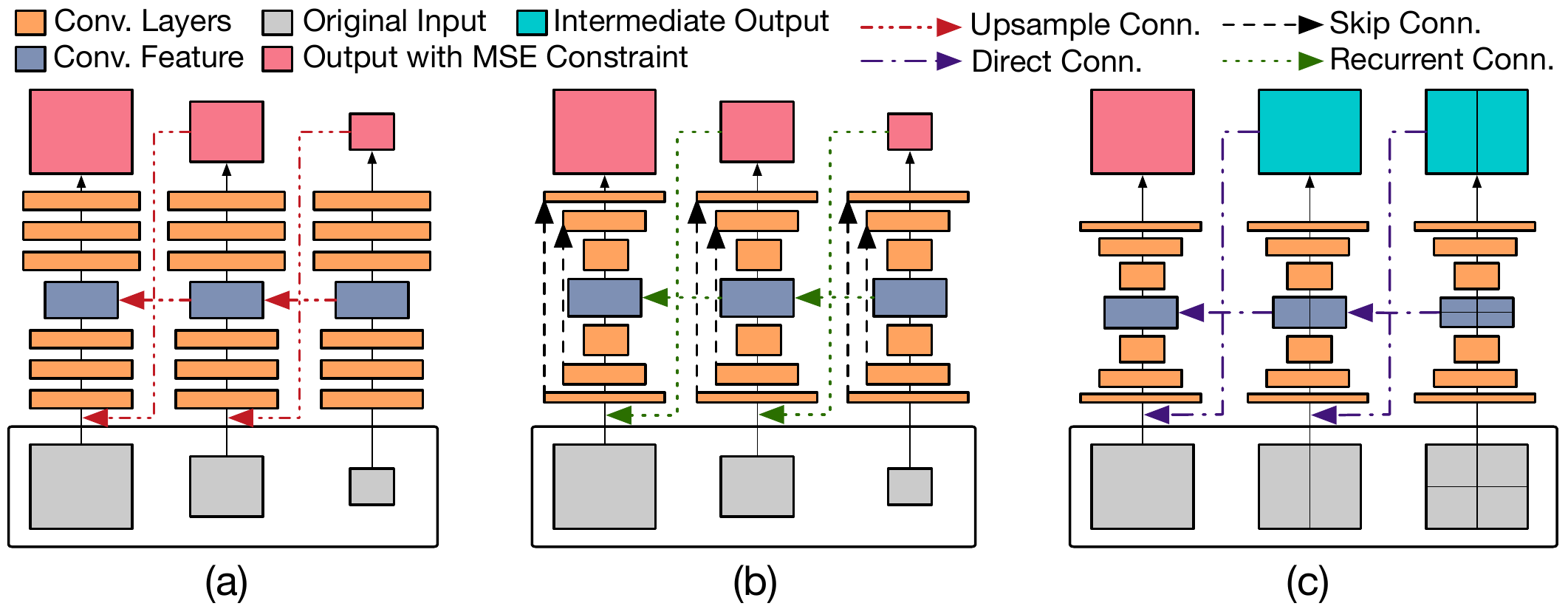}
	\caption{\small Comparison between different network architectures (a) multi-scale \cite{nah2017deep}, (b) scale-recurrent \cite{tao2018scale} and (c) our hierarchical multi-patch architecture. We do not employ any skip or recurrent connections which simplifies our model. Best viewed in color.}
	\label{fig:Structure_Comp}
    \vspace{-0.3cm}
\end{figure}
\fi

The advantages of our network are twofold: 1) As the inputs at different levels have the same spatial resolution, we can apply residual-like learning which requires small filter sizes and leads to a fast  inference; 2) We use an SPM-like  model which is exposed to more training data at the finest level due to relatively more patches available for that level.

In addition, we have observed a limitation to {\em stacking depth} on  multi-scale and multi-patch models, thus increasing the model depth by introducing additional coarser or finer grids cannot improve the overall deblurring performance of known models. To address this issue, we present two stacked versions of our DMPHN, whose performance is higher compared to current state-of-the-art deblurring methods.
Our contributions are summarized below:
\renewcommand{\labelenumi}{\Roman{enumi}.}
\hspace{-1.0cm}
\begin{enumerate}[leftmargin=0.6cm]
\item We propose an end-to-end CNN hierarchical model 
akin to Spatial Pyramid Matching (SPM) that performs deblurring in the fine-to-coarse grids thus exploiting multi-patch localized-to-coarse operations. Each finer level acts in the residual manner by contributing its residual image to the coarser level thus allowing each level of network focus on different scales of blur.
\vspace{-1mm}
\item We identify the limitation to stacking depth of current deep deblurring models and introduce novel stacking approaches which overcome this limitation.
\vspace{-1mm}
\item We perform  baseline comparisons in the common testbed (where possible) for  fair comparisons.
\vspace{-2mm}
\item We investigate the influence of weight sharing between the encoder-decoder pairs across hierarchy levels, and we propose a memory-friendly variant of DMPHN.
\vspace{-1mm}
\end{enumerate}

Our experiments will demonstrate clear benefits of our SPM-like model in motion deblurring. To the best of our knowledge, our CNN model  is the first multi-patch take on blind motion deblurring and DMPHN is the first model that supports deblurring of 720p images real-time (at 30fps). 
\section{Related Work}
Conventional image deblurring methods \cite{cho2009fast,jia2007single,xu2010two,levin2007blind,rajagopalan2014motion,jia2014mathematical,hyun2015generalized,sellent2016stereo} fail to remove  non-uniform motion blur due to the use of spatially-invariant deblurring kernel. Moreover, their complex computational inference leads to long processing times, which cannot satisfy the ever-growing needs for real-time  deblurring.

\noindent\textbf{Deep Deblurring.} Recently, CNNs have been used in non-uniform image deblurring to deal with the complex motion blur in a time-efficient manner \cite{xu2014deep,sun2015learning,nah2017deep,schuler2016learning,nimisha2017blur,su2017deep}.
Xu \etal \cite{xu2014deep} proposed a deconvolutional CNN which removes blur in non-blind setting by recovering a sharp image given the estimated blur kernel. Their network uses separable kernels which can be decomposed into a small set of filters.  
Sun \etal \cite{sun2015learning} estimated and removed a non-uniform motion blur from an image by learning the regression between 30$\times$30 image patches and their corresponding kernels. Subsequently, the conventional energy-based optimization is employed to estimate the latent sharp image. 

Su \etal \cite{su2017deep} presented a deep learning framework to process blurry video sequences and accumulate information across frames. This method does not require spatially-aligned pairs of samples. 
Nah \etal \cite{nah2017deep} exploited a multi-scale CNN to restore sharp images in an end-to-end fashion from images whose blur is caused by various factors. A multi-scale loss function is employed to mimic the coarse-to-fine pipeline in conventional deblurring approaches.

Recurrent Neural Network (RNN) is a popular tool employed in deblurring due to its advantage in sequential information processing. A network consisting of three deep CNNs and one RNN, proposed by \cite{zhang2018dynamic}, is a prominent example. The RNN is applied as a deconvolutional decoder on feature maps extracted by the first CNN module. Another CNN module learns weights for each layer of RNN. The last CNN module reconstructs the sharp image from deblurred feature maps. Scale-Recurrent Network (SRN-DeblurNet) \cite{tao2018scale} uses ConvLSTM cells to aggregate feature maps from coarse-to-fine scales. This shows the advantage of RNN units in non-uniform image deblurring task.

Generative Adversarial Nets (GANs) have also been employed in deblurring due to their advantage in preserving texture details and generating photorealistic images. Kupyn \etal \cite{kupyn2017deblurgan} presented a conditional GAN  which produces high-quality delburred images via the Wasserstein loss.

\vspace{-0.1cm}
\section{Our Framework}

In this paper, we propose to exploit the multi-patch hierarchy for efficient and effective blind motion deblurring. The overall architecture of our proposed DMPHN network is shown in Fig.~\ref{fig:DMPHN} fro which we use the (1-2-4-8) model (explained in Sec. \ref{sec:na}) as an example.  
Our network is inspired by coarse-to-fine Spatial Pyramid Matching \cite{lazebnik2006beyond}, which has been used for scene recognition \cite{koniusz2018deeper} to aggregate multiple image patches for better performance. In contrast to the expensive inference in multi-scale and scale-recurrent network models, our approach uses residual-like architecture, thus requiring small-size filters which result in fast processing. 
The differences between \cite{nah2017deep,tao2018scale} and our network architecture are illustrated in Fig.~\ref{fig:Structure_Comp}. Despite our model uses a very simple architecture (skip and recurrent connections have been removed), it is very effective. In contrast to \cite{nah2017deep} which uses deconvolution/upsampling links, we use operations such as feature map concatenations, which are possible due to the multi-patch setup we propose. 

\subsection{Encoder-decoder Architecture}
\begin{figure}[t]
	\centering
	\includegraphics[width=1.05\linewidth]{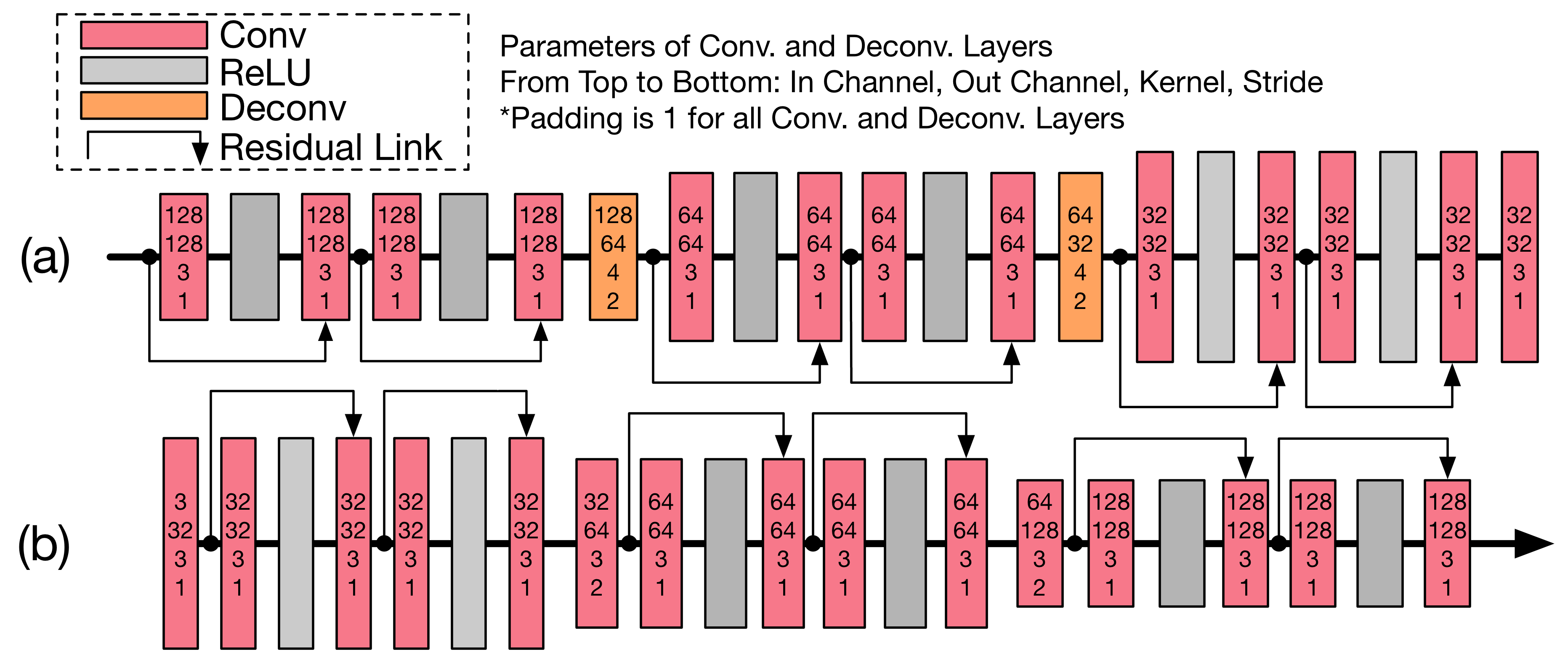}
	\caption{\small The architecture and layer configurations of our (a) decoder and (b) encoder.}
	\label{fig:encoder}
    \vspace{-0.4cm}
\end{figure}

Each level of our DMPHN network consists of one encoder and one decoder whose architecture is illustrated in Fig.~\ref{fig:encoder}.
Our encoder consists of 15 convolutional layers, 6 residual links and 6 ReLU units. The layers of decoder and encoder are identical except that two convolutional layers are replaced by deconvolutional layers to generate images.

The parameters of our encoder and decoder amount to 3.6 MB due to residual nature of our model which contributes significantly to the fast deblurring runtime. By contrast, the multi-scale deblurring network in \cite{nah2017deep} has 303.6 Mb parameters which results in the expensive inference. 

\hspace{-0.5cm}

\begin{figure*}[h]
\vspace{-0.8cm}
	\centering
	\includegraphics[width=0.95\linewidth]{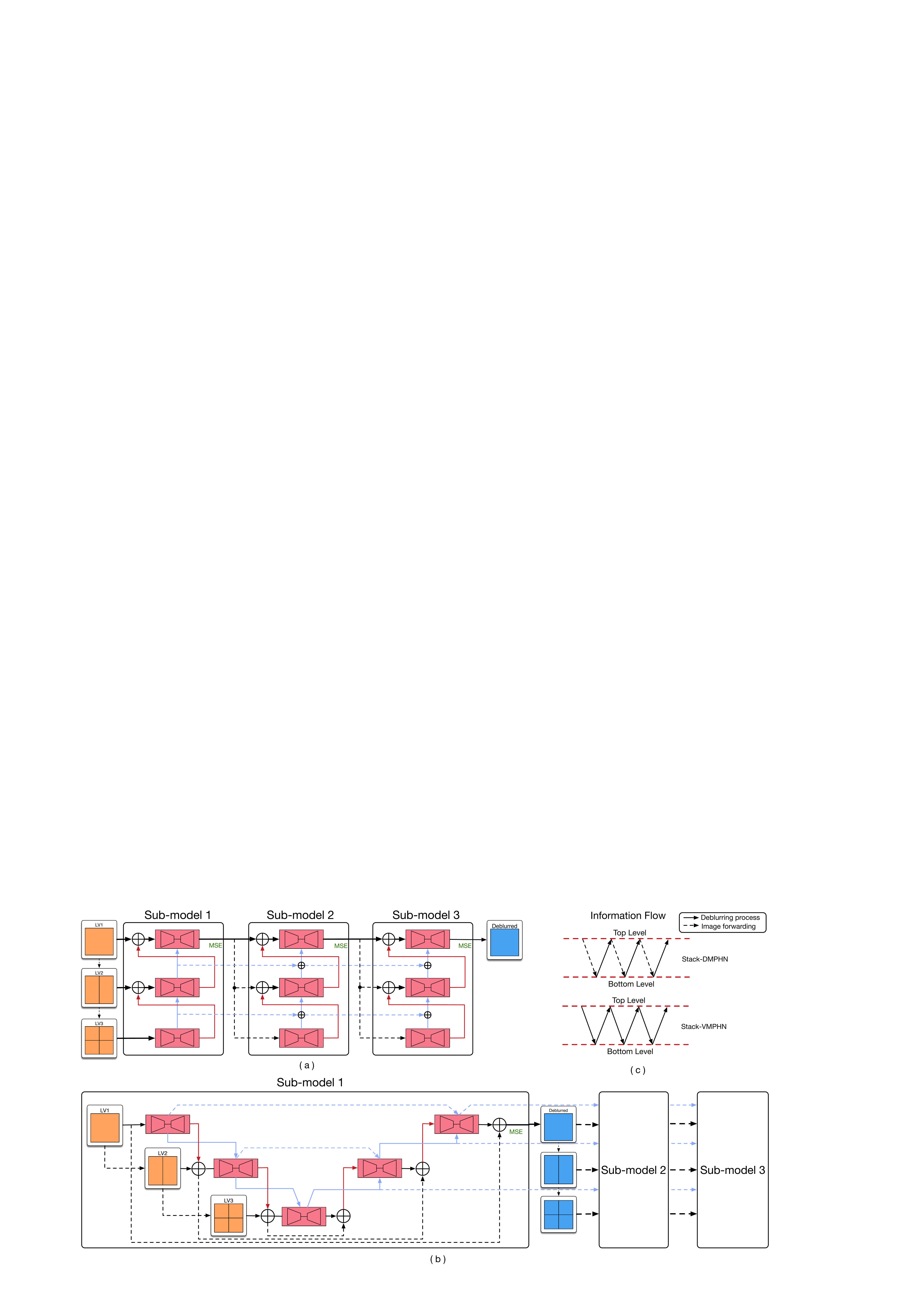}
	\caption{\small The architecture of stacking network. (a) Stack-DMPHN. (b)  Stack-VMPHN. (c) The information flow for two different stacking approaches. Note that the units in both stacking networks have (1-2-4) multi-patch hierarchical architecture. The model size of VMPHN unit is 2$\times$ as large as DMPHN unit.}
	\label{fig:stack}
    \vspace{-0.4cm}
\end{figure*}

\subsection{Network Architecture}
\label{sec:na}
The overall architecture of our DMPHN network is depicted in Fig.~\ref{fig:DMPHN}, in which we use the (1-2-4-8) model for illustration purposes.  
Notation (1-2-4-8) indicates the numbers of image patches from the coarsest to the finniest level \ie, a vertical split at the second level, $2\times 2 = 4$ splits at the third level, and $2\times 4 = 8$ splits at the fourth level.

We denote the initial blurry image input as ${\mathbf B}_1$, while ${\mathbf B}_{i，j}$ is the $j$-th patch at the $i$-th level. Moreover, $\mathcal{F}_i$ and $\mathcal{G}_i$ are the encoder and decoder at level $i$, ${\mathbf C}_{ij}$ is the output of $\mathcal{G}_i$ for ${\mathbf B}_{ij}$, and ${\mathbf S}_{ij}$ represents the output patches from $\mathcal{G}_i$.

Each level of our network consists of an encoder-decoder pair. The input for each level is generated by dividing the original blurry image input ${\mathbf B}_1$ into multiple non-overlapping patches. The output of both encoder and decoder from lower level (corresponds to finer grid) will be added to the upper level (one level above) so that the top level contains all information inferred in the finer levels. Note that the numbers of input and output patches at each level are different as the main idea of our work is to make the lower level focus on local information (finer grid) to produce residual information for the coarser gird (obtained by concatenating convolutional features).

Consider the (1-2-4-8) variant as an example. The deblurring process of DMPHN starts at the bottom level 4. ${\mathbf B}_1$ is sliced into 8 non-overlapping patches ${\mathbf B}_{4,j}, j\!=\!1,\cdots,8$, which are fed into the encoder $\mathcal{F}_4$ to produce the following convolutional feature representation:
\vspace{-0.1cm}
\begin{equation}
{\mathbf C}_{4,j} = \mathcal{F}_4({\mathbf B}_{4,j}), \quad j\in\{1...8\}.
\end{equation}

Then, we concatenate adjacent features (in the spatial sense) to obtain a new feat. representation $C^*_{4,j}$, which is of the same size as the conv. feat. representation at level 3:
\vspace{-0.1cm}
\begin{equation}
{\mathbf C}^*_{4,j} = {\mathbf C}_{4,2j-1} \oplus {\mathbf C}_{4,2j}, \quad j\in\{1...4\},
\end{equation}
where $\oplus$ denotes the concatenation operator. The concatenated feature representation ${\mathbf C}^*_{4,j}$ is passed through the encoder $\mathcal{G}_4$ to produce ${\mathbf S}_{4,j} = \mathcal{G}_4({\mathbf C}^*_{4,j})$. 

Next, we move one level up to level 3. The input of $\mathcal{F}_3$ is formed by summing up ${\mathbf S}_{4,j}$ with the sliced patches ${\mathbf B}_{3,j}$. Once the output of $\mathcal{F}_3$ is produced, we add to it ${\mathbf C}^*_{4,j}$:
\vspace{-0.2cm}
\begin{equation}
{\mathbf C}_{3,j} = \mathcal{F}_3({\mathbf B}_{3,j} + {\mathbf S}_{4,j}) + {\mathbf C}^*_{4,j}, \quad j\in\{1...4\}.
\end{equation}
\vspace{-0.2cm}

At level 3, we concatenate the feature representation of level 3 to obtain ${\mathbf C}^*_{3,j}$ and pass it through $\mathcal{G}_3$ to obtain ${\mathbf S}_{3,j}$:
\begin{align}
&{\mathbf C}^*_{3,j} = {\mathbf C}_{3,2j-1} \oplus {\mathbf C}_{3,2j}, \quad j\in\{1, 2\},\\
& {\mathbf S}_{3,j} = \mathcal{G}_3({\mathbf C}^*_{3,j}), \quad j\in\{1,2\}.
\end{align}

Note that features at all levels are concatenated along spatial dimension: imagine neighboring patches being concatenated to form a larger ``image''.

At level 2, our network takes two image patches ${\mathbf B}_{2,1}$ and ${\mathbf B}_{2,2}$ as input. We update ${\mathbf B}_{2,j}$ so that ${\mathbf B}_{2,j}\!:=\!{\mathbf B}_{2,j} + {\mathbf S}_{3,j}$ and pass it through $\mathcal{F}_2$:
\vspace{-0.1cm}
\begin{align}
&{\mathbf C}_{2,j} = \mathcal{F}_2({\mathbf B}_{2,j} + {\mathbf S}_{3,j}) + {\mathbf C}^*_{3,j}, \quad j\in \{1,2\},\\
&{\mathbf C}^*_2 = {\mathbf C}_{2,1} \oplus {\mathbf C}_{2,2}.
\end{align}

The residual map at level 2 is given by:
\vspace{-0.1cm}
\begin{equation}
{\mathbf S}_2 = \mathcal{G}_2({\mathbf C}^*_2).
\end{equation}

At level 1, the final deblurred output $S_1$ is given by:
\vspace{-0.1cm}
\begin{align}
&{\mathbf C}_1 = \mathcal{F}_1({\mathbf B}_1 + {\mathbf S}_2) + {\mathbf C}^*_2,
&{\mathbf S}_1 = \mathcal{G}_1({\mathbf C}_1).
\end{align}

Different from approaches \cite{nah2017deep,tao2018scale} that evaluate the Mean Square Error (MSE) loss at each level, we evaluate the MSE loss only at the output of level 1 (which resembles res. network). The loss function of DMPHN is given as:
\begin{equation}
{\mathcal L} = \frac{1}{2} \|{\mathbf S}_1 - {\mathbf G} \|_F^2,
\end{equation}
where ${\mathbf G}$ denotes the ground-truth sharp image.
Due to the hierarchical multi-patch architecture, our network follows the principle of residual learning: the intermediate outputs at different levels ${\mathbf S}_i$ capture  image statistics at different scales. Thus, we evaluate the loss function only at level 1. We have investigated the use of multi-level MSE loss, which forces the outputs at each level to be close to the ground truth image. However, as expected, there is no visible performance gain achieved by using the multi-scale loss.

\begin{figure*}[t]
\vspace{-0.6cm}
	\centering
	\includegraphics[width=0.85\linewidth]{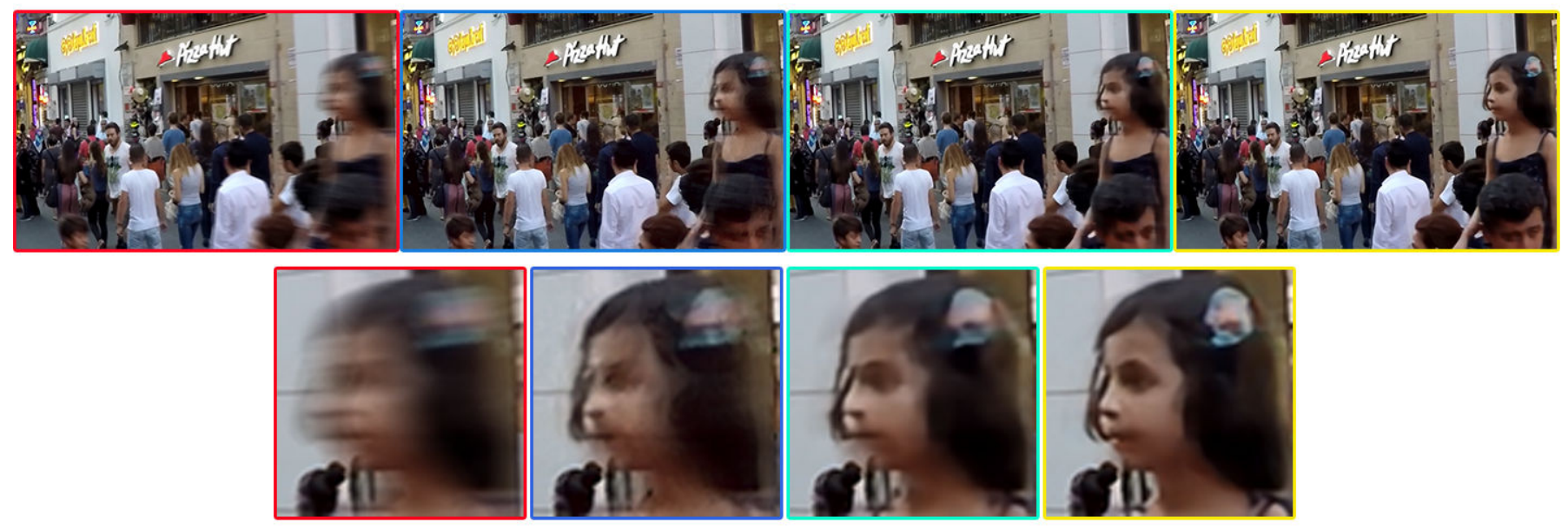}
	\caption{\small Deblurring results. Red block contains the blurred subject, blue and green are the results for \cite{nah2017deep} and \cite{tao2018scale}, respectively, yellow block indicates our result. As can be seen, our method produces the sharpest and most realistic facial details.}
	\label{fig:Comp1}
    \vspace{-0.4cm}
\end{figure*}

\vspace{-0.1cm}
\subsection{Stacked Multi-Patch Network}
As reported by Nah \etal \cite{nah2017deep} and Tao \etal \cite{tao2018scale}, adding finer network levels cannot improve the deblurring performance of the multi-scale and scale-recurrent architectures. For our multi-patch network, we have also observed that dividing the blurred image into ever smaller grids does not further improve the deblurring performance. 
This is mainly due to coarser levels attaining low empirical loss on the training data fast thus excluding the finest levels from contributing their residuals.

In this section, we propose a novel stacking paradigm for  deblurring. Instead of making the network deeper vertically (adding finer levels into the network model, which increases the difficulty for a single worker), we propose to increase the depth horizontally (stacking multiple network models), which employs multiple workers (DMPHN) horizontally to perform deblurring. 

Network models can be cascaded in numerous ways. In Fig.~\ref{fig:stack}, we provide two diagrams to demonstrate the proposed models. The first model, called Stack-DMPHN, stacks multiple ``bottom-top'' DMPHNs as shown in Fig.~\ref{fig:stack} (top). Note that the output of sub-model $i-1$ and the input of sub-model $i$ are connected, which means that for the optimization of sub-model $i$, output from the sub-model $i-1$ is required. All intermediate features of sub-model $i-1$ are passed to sub-model $i$. The MSE loss is evaluated at the output of every sub-model $i$. 

Moreover, we investigate a reversed direction of information flow, and propose a Stacked v-shape ``top-bottom-top'' multi-patch hierarchical network (Stack-VMPHN). We will show in our experiments that the Stack-VMPHN  outperforms DMPHN. The architecture of Stack-VMPHN is shown in Fig.~\ref{fig:stack} (bottom). We evaluate the MSE loss at the output of each sub-model of Stack-VMPHN. 

The Stack-VMPHN is built from our basic DMPHN units and it can be regarded as a reversed version of Stack(2)-DMPHN (2 stands for stacking of two sub-models). In Stack-DMPHN,  processing starts from the bottom level and ends at the top-level, then the output of the top-level is forwarded to the bottom level of next model. However, VMPHN begins from the top level, reaches the bottom level, and then it proceeds back to the top level.

The objective to minimize for both Stack-DMPHN and Stack-VMPHN is simply given as:
\vspace{-0.1cm}
\begin{equation}
    {\mathcal L} = \frac{1}{2}\sum\limits_{i=1}^N  \| {\mathbf S}_i - {\mathbf G}\|_F^2,
    \vspace{-0.1cm}
\end{equation}
where $N$ is the number of sub-models used, ${\mathbf S}_i$ is the output of sub-model $i$, and ${\mathbf G}$ is the ground-truth sharp image. 

Our experiments will illustrate that these two stacked networks improve the deblurring performance. Although our stacked architectures use  DMPHN units, we believe they are generic frameworks--other deep deblurring methods can be stacked in the similar manner to improve their performance. However, the total processing time may be unacceptable if a costly deblurring model is employed for the basic unit. Thanks to fast and efficient DMPHN units, we can control the runtime and size of stacking networks within a reasonable range to address various applications. 

\vspace{-0.1cm}
\subsection{Network Visualization}
\begin{figure}[h]
	\vspace{-0.2cm}
	\centering
	\includegraphics[width=\linewidth]{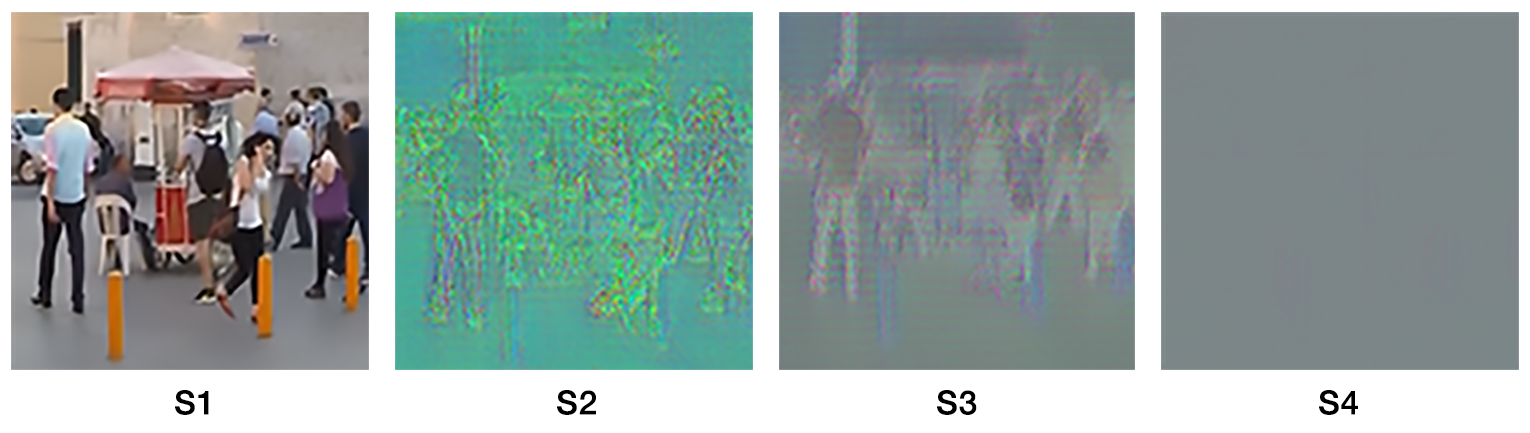}
    \caption{\small Outputs ${\mathbf S}_i$ for different levels of DMHPN(1-2-4-8). Images from right to left visualize bottom level ${\mathbf S}_4$ to top level ${\mathbf S}_1$.}
	\label{fig:dmphn_comp}
    \vspace{-0.2cm}
\end{figure}

We visualize the outputs of our DMPHN unit in Fig.~\ref{fig:dmphn_comp} to analyze its intermediate contributions.
As previously explained, DMPHN uses the residual design. Thus, finer levels contain finer but visually less important information compared to the coarser levels. In Fig.~\ref{fig:dmphn_comp}, we illustrate outputs ${\mathbf S}_i$ of each level of DMPHN (1-2-4-8). The information contained in ${\mathbf S}_4$ is the finest and most sparse. The outputs become less sparse, sharper and richer in color as we move up level-by-level.

\begin{figure}[h]
	\vspace{-0.2cm}
	\centering
	\includegraphics[width=\linewidth]{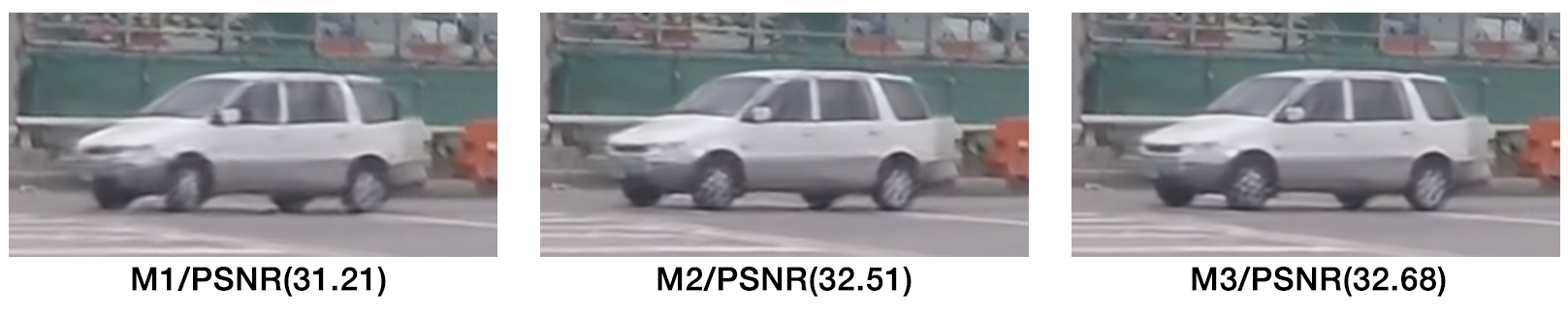}
    \caption{\small Outputs of different sub-models of Stack(3)-DMHPN. From left to right are the outputs of ${\mathbf M}_1$ to ${\mathbf M}_3$. The clarity of results improves level-by-level. We observed the similar behavior for Stack-VMPHN (not shown for brevity).}
	\label{fig:sdnet_comp}
    \vspace{-0.4cm}
\end{figure}

For the stacked model, the output of every sub-model is optimized level-by-level, which means the first output has the poorest quality and the last output achieves the best performance. Fig.~\ref{fig:sdnet_comp} presents the outputs of Stack(3)-DMPHN (3 sub-models stacked together) to demonstrate that each sub-model gradually improves the quality of deblurring.

\subsection{Implementation Details}
All our experiments are implemented in PyTorch and evaluated on a single NVIDIA Tesla P100 GPU. To train DMPHN, we randomly crop images to $256\!\times\!256$ pixel size. Subsequently, we extract patches from the cropped images  and forward them to the inputs of each level.
The batch size is set to 6 during training. The Adam solver \cite{kingma2014adam} is used to train our models for 3000 epochs. The initial learning rate is set to 0.0001 and the decay rate to 0.1. 
We normalize image to range $[0, 1]$ and subtract 0.5. 

\begin{table}[h]
\caption{\small Quantitative analysis of our model on the GoPro dataset \cite{nah2017deep}. {\em Size} and {\em Runtime} are expressed in MB and seconds. The reported time is the CNN runtime (writing generated images to disk is not considered). Note that we employ (1-2-4) hierarchical unit for both Stack-DMPHN and Stack-VMPHN. We did not investigate deeper stacking networks due to the GPU memory limits and long training times.}
\makebox{\begin{tabular}{|l|c|c|c|c|}
\hline
Models & \small PSNR & \small SSIM & Size & \small Runtime\\ \hline
Sun \etal \cite{sun2015learning} & 24.64 & 0.8429 & 54.1 & {\small 12000} \\
Nah \etal \cite{nah2017deep} & 29.23 & 0.9162 & 303.6 & {\small 4300}\\
Zhang \etal \cite{zhang2018dynamic} & 29.19 & 0.9306 & 37.1 & {\small 1400}\\
Tao \etal \cite{tao2018scale} & 30.10 & 0.9323 & 33.6 & {\small 1600}\\ \hline
\small DMPHN(1) & 28.70 &  0.9131 & 7.2 & 5\\
\small DMPHN(1-2) & 29.77 & 0.9286 & 14.5 & 9\\
\small DMPHN(1-1-1) & 28.11 & 0.9041 & 21.7 & 12\\
\small DMPHN(1-2-4) & 30.21 & 0.9345 & 21.7 & 17\\
\small DMPHN(1-4-16) & 29.15 & 0.9217 & 21.7 & 92\\
\small DMPHN(1-2-4-8) & \textbf{30.25} & \textbf{0.9351} & 29.0 & 30\\
\small DMPHN(1-2-4-8-16) & 29.87 & 0.9305 & 36.2 & 101 \\
\hline
\small DMPHN & 30.21 & 0.9345 & 21.7 & 17\\
Stack(2)-DMPHN & 30.71 & 0.9403 & 43.4 & 37\\
Stack(3)-DMPHN & 31.16 & 0,9451 & 65.1 & 233\\
Stack(4)-DMPHN & \textbf{31.20} & \textbf{0.9453} & 86.8 & 424 \\
\hline
VMPHN & 30.90 & 0.9419 & 43.4 & 161\\
Stack(2)-VMPHN & \textbf{31.50} & \textbf{0.9483} & 86.8 & 552 \\
\hline
\end{tabular}}
\label{tabel:PSNR}
\vspace{-0.2cm}
\end{table}

\begin{table}[h]
\centering
\caption{\small The baseline performance of multi-scale and multi-patch methods on the GoPro dataset \cite{nah2017deep}. Note that DMSN(1) and DMPHN(1) are in fact the same model.} 
\makebox{\begin{tabular}{|l|c|c|c|}
\hline
Models & PSNR & SSIM & Runtime\\ \hline
Nah \etal \cite{nah2017deep} & 29.23 & 0.9162 & 4300\\ \hline
DMSN(1) &  \multirow{2}{*}{28.70} & \multirow{2}{*}{0.9131} & \multirow{2}{*}{4}\\ 
DMPHN(1) &    &  & \\
\arrayrulecolor{black} \cdashline{1-4}[1pt/3pt]
DMSN(2) &  28.82 & 0.9156 & 21\\
DMPHN(1-2) & 29.77 & 0.9286 & 9 \\
\arrayrulecolor{black} \cdashline{1-4}[1pt/3pt]
DMSN(3) &  28.97 &  0.9178 & 27\\
DMPHN(1-2-4) & 30.21 & 0.9345 & 17\\
\hline
\end{tabular}}
\label{tabel:PSNR_sc}
\vspace{-0.3cm}
\end{table}

\begin{table*}[t]
\vspace{-0.8cm}
\centering
\caption{\small Quantitative analysis (PSNR) on the VideoDeblurring dataset \cite{su2017deep} for models trained on GoPro dataset. PSDeblur means using Photoshop CC 2015.
We select the ``single frame'' version of approach \cite{su2017deep} for fair comparisons.}
\makebox{\begin{tabular}{|l|c|c|c|c|c|c|c|c|c|c|c|c|c|}
\hline
Methods & \#1 & \#2 & \#3 & \#4 & \#5 & \#6 & \#7 & \#8 & \#9 & \#10 & {\small Average} \\ \hline
Input  & 24.14 & 30.52 & 28.38 & 27.31 & 22.60 & 29.31 & 27.74 & 23.86 & 30.59 & 26.98 & 27.14 \\
PS\small{Deblur}  &24.42 & 28.77 & 25.15 & 27.77 & 22.02 & 25.74 & 26.11 & 19.75 & 26.48 & 24.62 & 25.08 \\
WFA \cite{delbracio2015hand} & 25.89 & 32.33 & 28.97 & 28.36 & 23.99 & 31.09 & 28.58 & 24.78 & 31.30 & 28.20 & 28.35 \\
Su \etal \cite{su2017deep} & 25.75 & 31.15 & 29.30 & 28.38 & 23.63 & 30.70 & 29.23 & 25.62 & 31.92 & 28.06 & 28.37 \\ \hline
DMPHN & 29.89 & 33.35 & 31.82 & 31.32 & 26.35 & 32.49 & 30.51 & 27.11 & 34.77 & 30.02 & 30.76 \\
Stack(2)-DMPHN & 30.19 & 33.98 & 32.16 & 31.82 & 26.57 & 32.94 & 30.73 & 27.45 & 35.11 & 30.41 & 31.22 \\
Stack(3)-DMPHN & 30.48 & 34.31 & 32.24 & 32.09 & 26.77 & 33.08 & 30.84 & 27.51 & 35.24 & 30.57 & 31.39 \\
Stack(4)-DMPHN & \textbf{30.48} & \textbf{34.41} & \textbf{32.25} & \textbf{32.10} & \textbf{26.87} & \textbf{33.12} & \textbf{30.86} & \textbf{27.55} & \textbf{35.25} & \textbf{30.60} & \textbf{31.43} \\\hline
\end{tabular}}
\label{tabel:videodeblurringresults}
\vspace{-0.3cm}
\end{table*}

\begin{table}[h]
\centering
\caption{\small Quantitative results for the weight sharing on GoPro \cite{nah2017deep}.}
\makebox{\begin{tabular}{|l|c|c|c|}
\hline
Models & PSNR & SSIM & Size (MB)\\ \hline
DMPHN(1-2) & 29.77 & 0.9286 & 14.5\\
DMPHN(1-2)-WS & 29.22 & 0.9210 & 7.2\\
\arrayrulecolor{black} \cdashline{1-4}[1pt/3pt]
DMPHN(1-2-4) & 30.21 & 0.9343 & 21.7\\
DMPHN(1-2-4)-WS & 29.56 & 0.9257 & 7.2\\
\arrayrulecolor{black} \cdashline{1-4}[1pt/3pt]
DMPHN(1-2-4-8) & 30.25 & 0.9351 & 29.0\\
DMPHN(1-2-4-8)-WS & 30.04 & 0.9318 & 7.2\\ \hline
\end{tabular}}
\label{tabel:share_weight}
\vspace{-0.4cm}
\end{table}

\vspace{-0.2cm}
\section{Experiments}
\subsection{Dataset}
We train/evaluate our methods on several versions of the GoPro dataset \cite{nah2017deep} and the VideoDeblurring dataset \cite{su2017deep}.

\vspace{0.05cm}
\noindent{\textbf{GoPro dataset}} \cite{nah2017deep} consists of 3214 pairs of blurred and clean images extracted from 33 sequences captured at 720$\times$1280 resolution. The blurred images  are generated by averaging varying number (7--13) of successive latent frames to produce varied blur. For a fair comparison, we follow the protocol in \cite{nah2017deep}, which uses 2103 image pairs for training and the remaining 1111 pairs for testing.

\noindent{\textbf{VideoDeblurring dataset}} \cite{su2017deep} contains videos captured by various devices, such as iPhone, GoPro and Nexus. The quantitative part has 71 videos. Every video consists of 100 frames at 720$\times$1280 resolution. Following the setup in \cite{su2017deep}, we use 61 videos for training and the remaining 10 videos for testing. In addition, we evaluate the model trained on the GoPro dataset\cite{nah2017deep} on the VideoDeblurring dataset to evaluate the generalization ability of our methods.

\subsection{Evaluation Setup and Results}
We feed the original high-resolution $720\!\times\!1280$ pixel images into DMPHN for performance analysis. The PSNR, SSIM, model size and runtime are reported in Table \ref{tabel:PSNR} for an in-depth comparison with competing state-of-the-art motion deblurring models. For stacking networks, we employ (1-2-4) multi-patch hierarchy in every model unit considering the runtime and difficulty of training.

\noindent{\textbf{Performance.}} As illustrated in Table \ref{tabel:PSNR}, our proposed DMPHN outperforms other competing methods according to PSNR and SSIM, which demonstrates the superiority of non-uniform blur removal via the localized information our model uses. The deepest DMPHN we trained and evaluated is (1-2-4-8-16) due to the GPU memory limitation. The best performance is obtained with (1-2-4-8) model, for which PSNR and SSIM are higher compared to all current state-of-the-art models. Note that our model is simpler than other competing approaches \eg, we do not use recurrent units. We note that patches that are overly small (below 1/16 size) are not helpful in removing the motion blur.

Moreover, stacked variant Stack(4)-DMPHN outperformed shallower model DMPHN by 1\% PSNR, VMPHN outperformed DMPHN by 0.7\% PSNR while stacked variant Stack(2)-VMPHN outperformed shallower DMPHN by $\sim$1.3\% PSNR. SSIM scores indicate the same trend.

\begin{figure*}[htp]
	\centering
	\includegraphics[width=\linewidth]{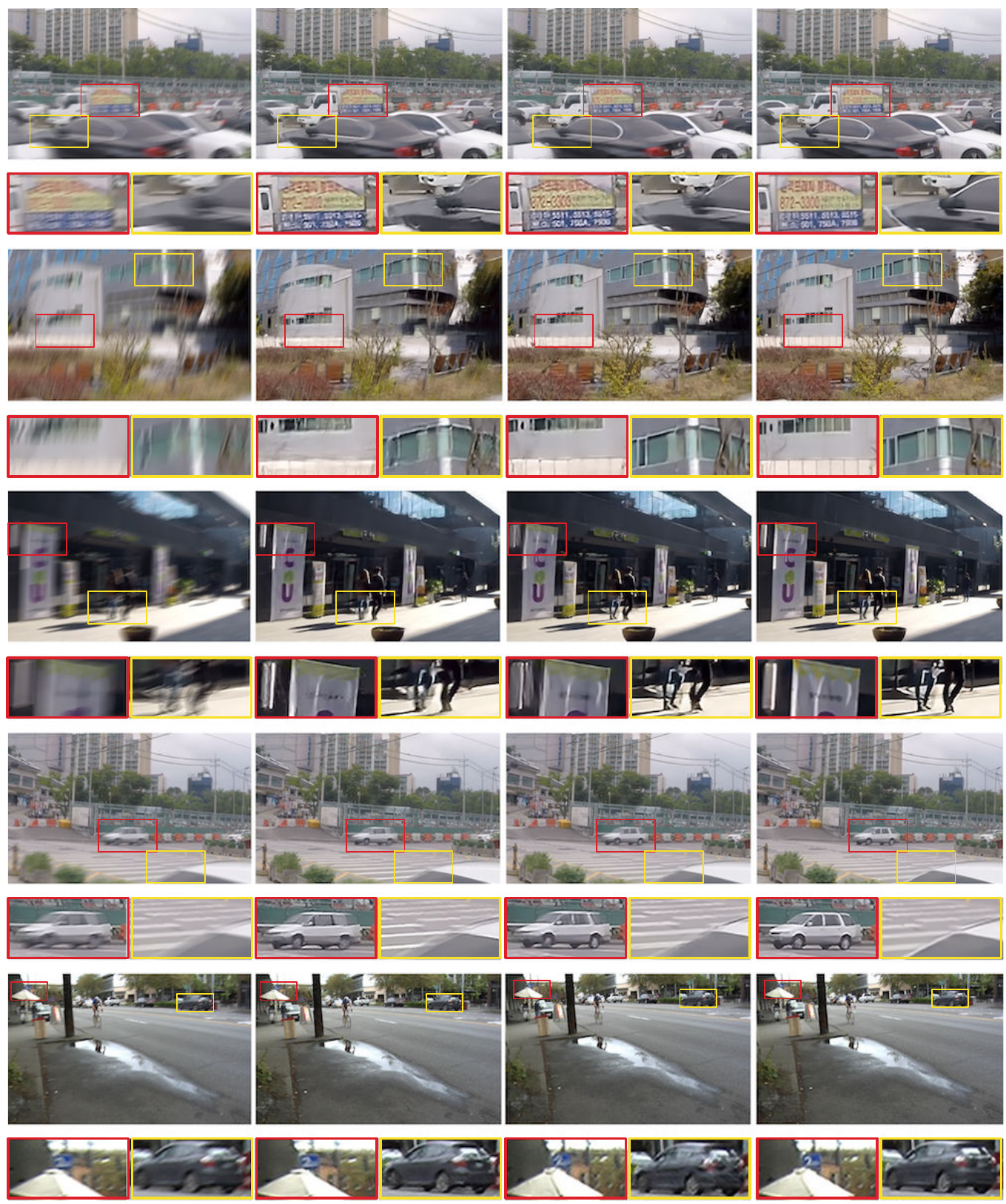}
	\caption{\small Deblurring performance on the blurry images from the GoPro and the VideoDeblurring datasets. The first column contains the original blurry images, the second column is the result of \cite{nah2017deep}, the third column is the result of \cite{tao2018scale}. Our results are presented in the last column. As can be seen, our model achieves the best performance across different scenes.}
	\label{fig:Comp2}
\end{figure*}

The deblurred images from the GoPro dataset are shown in Fig.~\ref{fig:Comp1} and \ref{fig:Comp2}. In Fig.~\ref{fig:Comp1}, we show the deblurring performance of different models for an image containing heavy motion blur. We zoom in the main object for clarity. In Fig.~\ref{fig:Comp2}, we select the images of different scenes to demonstrate the advantages of our model. As can be seen, our DMPHN produces the sharpest details in all cases.

\noindent\textbf{Runtime}. In addition to the superior PSNR and SSIM of our model, to the best of our knowledge, DMPHN is also the first deep deblurring model that can work in real-time. For example, DMPHN (1-2-4-8) takes 30ms to process a 720$\times$1280 image, which means it supports real-time 720p image deblurring at 30fps. However, there are runtime overheads related to I/O operations, so the real-time deblurring application requires fast transfers from a video grabber to GPU, larger GPU memory and/or an SSD drive, \etc. 

The following factors contribute to our fast runtime: i) shallower encoder-decoder with smal-size convolutional filters; ii) removal of unnecessary links \eg, skip or recurrent connections; iii) reduced number of upsampling/deconvolution between convolutional features of different levels. 

\noindent \textbf{Baseline Comparisons.} Although our model has a much better performance than the multi-scale model \cite{nah2017deep}, it is an unfair comparison as network architectures of our proposed model and \cite{nah2017deep} differ significantly. Compared with \cite{nah2017deep}, which uses over 303.6MB parameters, we apply much shallower CNN encoders and decoders with the model size 10$\times$ smaller. Thus, we create a Deep Multi-Scale Network (DMSN) that uses our encoder-decoder following the setup in \cite{nah2017deep} for the baseline comparison (sanity check) between multi-patch and multi-scale methods. As shown in Table \ref{tabel:PSNR_sc}, the PSNR of DMSN is worse than \cite{nah2017deep}, which is expected due to our simplified CNN architecture. Compared with our DMPHN, the best result obtained with DMSN is worse than the DMPHN(1-2) model. Due to the common testbed, we argue that the performance of DMSN and DMHPN reported by us is the fair comparison of the multi-patch hierarchical and multi-scale models \cite{nah2017deep}.

\vspace{0.05cm}
\noindent \textbf{Weight Sharing.} Below, we investigate weight sharing between the encoder-decoder pairs of all levels of our network to reduce the number of parameters in our model. Table \ref{tabel:share_weight} shows that weight sharing results in a slight loss of performance but reduces the number of parameters significantly.

\vspace{-0.2cm}
\section{Conclusions}
In this paper, we address the challenging problem of non-uniform motion deblurring by exploiting the multi-patch SPM- and residual-like model as opposed to the widely used multi-scale and scale-recurrent architectures. Based on oru proposition, we devised an end-to-end deep multi-patch hierarchical deblurring network. Compared against existing deep deblurring frameworks, our model achieves the state-of-the-art performance (according to PSNR and SSIM) and is able to run at 30fps for 720p images. Our work provides an insight for subsequent deep deblurring works regarding efficient deblurring.
Our stacked variants Stack(4)-DMPHN and Stack(2)-VMPHN further improved results over both shallower DMPHN and competing approaches while being $\sim\!\!4\!\times$ faster than the latter methods. Our stacking architecture appears to have overcome the limitation to stacking depth which other competing approaches exhibit. 

{
\vspace{0.1cm}
\noindent
\textbf{Acknowledgements.}
This research is supported in part by the Australian Research Council through Australian Centre for Robotic Vision (CE140100016), Australian Research Council grants (DE140100180), the China Scholarship Council (CSC Student ID 201603170283). Y. Dai is also funded in part by the Natural Science Foundation of China (61871325, 61420106007). Hongdong Li is also funded in part by ARC-DP (190102261) and ARC-LE (190100080). We also  thank for the support of CSIRO Scientific Computing, NVIDIA (GPU grant) and National University of Defense Technology. 
}

\begin{appendices}
\section{Outputs of Stacked Network}

\begin{figure}[h]
	\vspace{-0.1cm}
	\centering
	\includegraphics[width=\linewidth]{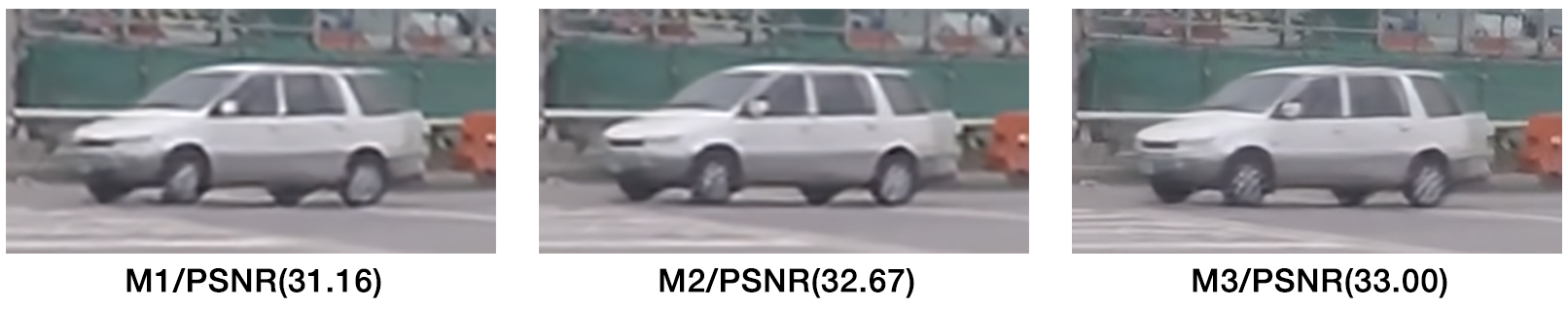}
    \caption{\small The outputs for different sub-models of Stack(3)-VMHPN. From left to right are the outputs of $\mathbf{M}_1$ to $\mathbf{M}_3$.}
	\label{fig:svdnet_comp}
    \vspace{-0.3cm}
\end{figure}

\begin{figure}[h]
	\vspace{-0.4cm}
	\centering
	\includegraphics[width=\linewidth]{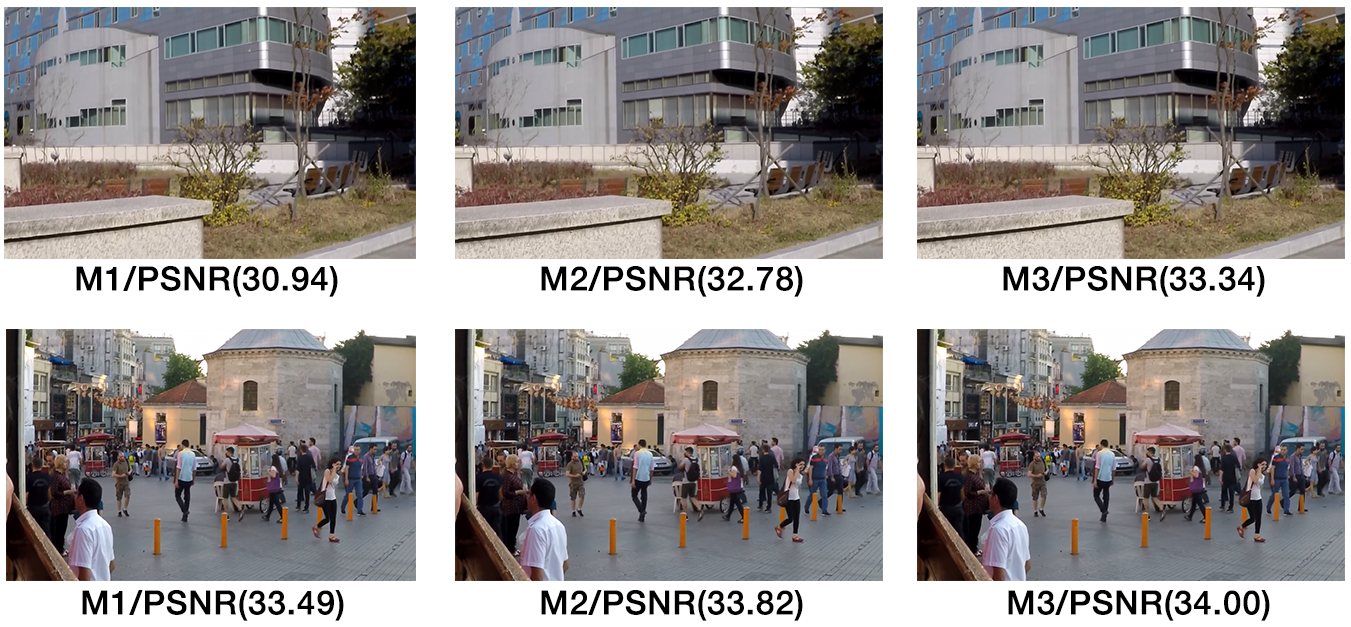}
    \caption{\small The outputs for different sub-models of Stack(3)-DMHPN. From left to right are the outputs of $\mathbf{M}_1$ to $\mathbf{M}_3$.}
	\label{fig:supp_comp}
    \vspace{-0.5cm}
\end{figure}

Below we present the intermediate outputs of our Stack-VMPHN. Figure \ref{fig:svdnet_comp} shows that the performance is optimized level by level, which is consistent with the behaviour of Stack-DMPHN. We also provide more instances for Stack-DMPHN to demonstrate its process in Figure \ref{fig:supp_comp}.

\section{Extension to Saliency Detection}
\begin{figure}[t]
    \vspace{-0.5cm}
    \centering
    \includegraphics[height=4.5cm]{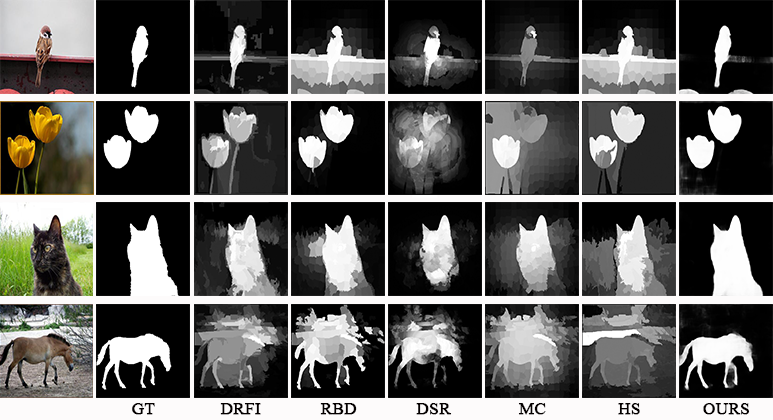}
    \caption{Instances of saliency detection on the MSRA-B dataset.}
    \label{fig:saliency_comp}
    \vspace{-0.3cm}
\end{figure}

We perform saliency detection with our proposed model to investigate the generalization ability on different tasks. Our proposed model is evaluated on the MSRA-B dataset. This dataset consists of 3000 images for training and 2000 images for testing. Note that all current deep methods of saliency detection highly depend on VGG or ResNet pre-trained on ImageNet and these methods often will not converge without pre-training on ImageNet. By contrast, our network can be easily trained from scratch. It outperforms all conventional methods and it is real-time. We evaluated single VMPHN for quantitative analysis. To make our network compatible with the saliency detection task, the output channel is modified to 1 for gray image generation, and the residual connection between input and output at level 1 is disabled in VMPHN. Figure \ref{fig:saliency_comp} and Table \ref{tabel:share_weight} show our results.

\begin{table}[h]
\caption{\small Quantitative analysis of saliency detection on MSRA-B. For $F_{\beta}$, higher scores are better. For MAE, lower scores are better.}
\makebox{\begin{tabular}{|l|c|c|c|c|c|c|}
\hline
\small Model & \small \cite{jiang2013salient}&\small \cite{zhu2014saliency} &\small \cite{li2013saliency} &\small \cite{jiang2013saliency} &\small \cite{zou2015harf} &\small OURS \\ \hline
\small $F_{\beta}$ & .728 & .751 & .723 & .717 & .713 & \textbf{.768}\\
\small{MAE} & .123 & .117 & .121 & .144 & .161 & \textbf{.107}\\ \hline
\end{tabular}}
\label{tabel:share_weight}
\vspace{-0.4cm}
\end{table}
\end{appendices}

{
\bibliographystyle{ieee_fullname}
\bibliography{Deblur-Reference}

\begin{thebibliography}{10}\itemsep=-1pt

\bibitem{cho2009fast}
Sunghyun Cho and Seungyong Lee.
\newblock Fast motion deblurring.
\newblock {\em ACM Transactions on graphics}, 28(5):145:1--145:8, 2009.

\bibitem{delbracio2015hand}
M. Delbracio and G. Sapiro.
\newblock Hand-held video deblurring via efficient fourier aggregation.
\newblock {\em IEEE Transactions on Computational Imaging}, 1(4):270--283, Dec
  2015.

\bibitem{he2014spatial}
Kaiming He, Xiangyu Zhang, Shaoqing Ren, and Jian Sun.
\newblock Spatial pyramid pooling in deep convolutional networks for visual
  recognition.
\newblock In {\em Proc. Eur. Conf. Comp. Vis.}, pages 346--361. Springer, 2014.

\bibitem{hyun2015generalized}
Tae Hyun~Kim and Kyoung Mu~Lee.
\newblock Generalized video deblurring for dynamic scenes.
\newblock In {\em Proc. IEEE Conf. Comp. Vis. Patt. Recogn.}, pages 5426--5434,
  2015.

\bibitem{jia2007single}
Jiaya Jia.
\newblock Single image motion deblurring using transparency.
\newblock In {\em Proc. IEEE Conf. Comp. Vis. Patt. Recogn.}, pages 1--8. IEEE,
  2007.

\bibitem{jia2014mathematical}
Jiaya Jia.
\newblock Mathematical models and practical solvers for uniform motion
  deblurring., 2014.

\bibitem{jiang2013saliency}
Bowen Jiang, Lihe Zhang, Huchuan Lu, Chuan Yang, and Ming-Hsuan Yang.
\newblock Saliency detection via absorbing markov chain.
\newblock In {\em Proc. IEEE Int. Conf. Comp. Vis.}, pages 1665--1672, 2013.

\bibitem{jiang2013salient}
Huaizu Jiang, Jingdong Wang, Zejian Yuan, Yang Wu, Nanning Zheng, and Shipeng
  Li.
\newblock Salient object detection: A discriminative regional feature
  integration approach.
\newblock In {\em Proc. IEEE Conf. Comp. Vis. Patt. Recogn.}, pages 2083--2090,
  2013.

\bibitem{kingma2014adam}
Diederik~P Kingma and Jimmy Ba.
\newblock Adam: A method for stochastic optimization.
\newblock {\em arXiv preprint arXiv:1412.6980}, 2014.

\bibitem{koniusz2018deeper}
Piotr Koniusz, Hongguang Zhang, and Fatih Porikli.
\newblock A deeper look at power normalizations.
\newblock In {\em Proc. IEEE Conf. Comp. Vis. Patt. Recogn.}, pages 5774--5783,
  2018.

\bibitem{krizhevsky2012imagenet}
Alex Krizhevsky, Ilya Sutskever, and Geoffrey~E Hinton.
\newblock Imagenet classification with deep convolutional neural networks.
\newblock In {\em Proc. Adv. Neural Inf. Process. Syst.}, pages 1097--1105,
  2012.

\bibitem{kupyn2017deblurgan}
Orest Kupyn, Volodymyr Budzan, Mykola Mykhailych, Dmytro Mishkin, and Jiri
  Matas.
\newblock Deblurgan: Blind motion deblurring using conditional adversarial
  networks.
\newblock {\em arXiv preprint arXiv:1711.07064}, 2017.

\bibitem{lazebnik2006beyond}
Svetlana Lazebnik, Cordelia Schmid, and Jean Ponce.
\newblock Beyond bags of features: Spatial pyramid matching for recognizing
  natural scene categories.
\newblock In {\em Proc. IEEE Conf. Comp. Vis. Patt. Recogn.}, pages 2169--2178.
  IEEE, 2006.

\bibitem{levin2007blind}
Anat Levin.
\newblock Blind motion deblurring using image statistics.
\newblock In {\em Proc. Adv. Neural Inf. Process. Syst.}, pages 841--848, 2007.

\bibitem{li2013saliency}
Xiaohui Li, Huchuan Lu, Lihe Zhang, Xiang Ruan, and Ming-Hsuan Yang.
\newblock Saliency detection via dense and sparse reconstruction.
\newblock In {\em Proc. IEEE Int. Conf. Comp. Vis.}, pages 2976--2983, 2013.

\bibitem{lu2015deep}
Xin Lu, Zhe Lin, Xiaohui Shen, Radomir Mech, and James~Z Wang.
\newblock Deep multi-patch aggregation network for image style, aesthetics, and
  quality estimation.
\newblock In {\em Proc. IEEE Conf. Comp. Vis. Patt. Recogn.}, pages 990--998,
  2015.

\bibitem{nah2017deep}
Seungjun Nah, Tae~Hyun Kim, and Kyoung~Mu Lee.
\newblock Deep multi-scale convolutional neural network for dynamic scene
  deblurring.
\newblock In {\em Proc. IEEE Conf. Comp. Vis. Patt. Recogn.}, pages 257 -- 265,
  2017.

\bibitem{nimisha2017blur}
Thekke~Madam Nimisha, Akash~Kumar Singh, and A.~N. Rajagopalan.
\newblock Blur-invariant deep learning for blind-deblurring.
\newblock In {\em Proc. IEEE Int. Conf. Comp. Vis.}, pages 4762--4770, 2017.

\bibitem{pan2017simultaneous}
Liyuan Pan, Yuchao Dai, Miaomiao Liu, and Fatih Porikli.
\newblock Simultaneous stereo video deblurring and scene flow estimation.
\newblock In {\em Proc. IEEE Conf. Comp. Vis. Patt. Recogn.}, pages 6987--6996.
  IEEE, 2017.

\bibitem{rajagopalan2014motion}
A.~N. Rajagopalan and Rama Chellappa.
\newblock {\em Motion Deblurring: Algorithms and Systems}.
\newblock Cambridge University Press, 2014.

\bibitem{schuler2016learning}
Christian Schuler, Michael Hirsch, Stefan Harmeling, and Bernhard Scholkopf.
\newblock Learning to deblur.
\newblock {\em {IEEE} Trans. Pattern Anal. Mach. Intell.}, (7):1439--1451,
  2016.

\bibitem{sellent2016stereo}
Anita Sellent, Carsten Rother, and Stefan Roth.
\newblock Stereo video deblurring.
\newblock In {\em Proc. Eur. Conf. Comp. Vis.}, pages 558--575. Springer, 2016.

\bibitem{simonyan2014very}
Karen Simonyan and Andrew Zisserman.
\newblock Very deep convolutional networks for large-scale image recognition.
\newblock {\em arXiv preprint arXiv:1409.1556}, 2014.

\bibitem{su2017deep}
Shuochen Su, Mauricio Delbracio, Jue Wang, Guillermo Sapiro, Wolfgang Heidrich,
  and Oliver Wang.
\newblock Deep video deblurring for hand-held cameras.
\newblock In {\em Proc. IEEE Conf. Comp. Vis. Patt. Recogn.}, volume~2, pages
  237 -- 246, 2017.

\bibitem{sun2015learning}
Jian Sun, Wenfei Cao, Zongben Xu, and Jean Ponce.
\newblock Learning a convolutional neural network for non-uniform motion blur
  removal.
\newblock In {\em Proc. IEEE Conf. Comp. Vis. Patt. Recogn.}, pages 769--777,
  2015.

\bibitem{tao2018scale}
Xin Tao, Hongyun Gao, Xiaoyong Shen, Jue Wang, and Jiaya Jia.
\newblock Scale-recurrent network for deep image deblurring.
\newblock In {\em Proc. IEEE Conf. Comp. Vis. Patt. Recogn.}, pages 8174--8182,
  2018.

\bibitem{xu2010two}
Li Xu and Jiaya Jia.
\newblock Two-phase kernel estimation for robust motion deblurring.
\newblock In {\em Proc. Eur. Conf. Comp. Vis.}, pages 157--170. Springer, 2010.

\bibitem{xu2014deep}
Li Xu, Jimmy~SJ Ren, Ce Liu, and Jiaya Jia.
\newblock Deep convolutional neural network for image deconvolution.
\newblock In {\em Proc. Adv. Neural Inf. Process. Syst.}, pages 1790--1798,
  2014.

\bibitem{zhang2018dynamic}
Jiawei Zhang, Jinshan Pan, Jimmy Ren, Yibing Song, Linchao Bao, Rynson~WH Lau,
  and Ming-Hsuan Yang.
\newblock Dynamic scene deblurring using spatially variant recurrent neural
  networks.
\newblock In {\em Proc. IEEE Conf. Comp. Vis. Patt. Recogn.}, pages 2521--2529,
  2018.

\bibitem{zhu2014saliency}
Wangjiang Zhu, Shuang Liang, Yichen Wei, and Jian Sun.
\newblock Saliency optimization from robust background detection.
\newblock In {\em Proc. IEEE Conf. Comp. Vis. Patt. Recogn.}, pages 2814--2821,
  2014.

\bibitem{zou2015harf}
Wenbin Zou and Nikos Komodakis.
\newblock Harf: Hierarchy-associated rich features for salient object
  detection.
\newblock In {\em Proc. IEEE Int. Conf. Comp. Vis.}, pages 406--414, 2015.

\end{thebibliography}
}

\end{document}